\documentclass[letterpaper]{article} 
\usepackage{aaai24}  
\usepackage{times}  
\usepackage{helvet}  
\usepackage{courier}  
\usepackage[hyphens]{url}  
\usepackage{graphicx} 
\usepackage{natbib}  
\usepackage{caption} 
\usepackage{algorithm}
\usepackage{algorithmic}
\usepackage{amsmath}
\usepackage{multirow}
\usepackage{graphicx}
\usepackage{xcolor}
\usepackage{eqparbox}
\usepackage{subcaption}

 \usepackage{booktabs}

\newcommand{\floor}[1]{\left\lfloor #1 \right\rfloor}

\usepackage{newfloat}
\usepackage{listings}
\DeclareCaptionStyle{ruled}{labelfont=normalfont,labelsep=colon,strut=off} 
\floatstyle{ruled}
\newfloat{listing}{tb}{lst}{}
\floatname{listing}{Listing}
\nocopyright

\title{PARs: Predicate-based Association Rules for Efficient and Accurate Model-Agnostic Anomaly Explanation}

\author {
    Cheng Feng
}
\affiliations{
    Siemens Technology China\\
    cheng.feng@siemens.com
}

\usepackage{bibentry}

\begin{document}

\maketitle

\begin{abstract}
While new and effective methods for anomaly detection are frequently introduced, many studies prioritize the detection task without considering the need for explainability. Yet, in real-world applications, anomaly explanation, which aims to provide explanation of why specific data instances are identified as anomalies, is an equally important task. In this work, we present a novel approach for efficient and accurate model-agnostic anomaly explanation for tabular data using Predicate-based Association Rules (PARs). PARs can provide intuitive explanations not only about which features of the anomaly instance are abnormal, but also the reasons behind their abnormality. Our user study indicates that the anomaly explanation form of PARs is better comprehended and preferred by regular users of anomaly detection systems as compared to existing model-agnostic explanation options. Furthermore, we conduct extensive experiments on various benchmark datasets, demonstrating that PARs compare favorably to state-of-the-art model-agnostic methods in terms of computing efficiency and explanation accuracy on anomaly explanation tasks. The code for PARs tool is available at https://github.com/NSIBF/PARs-EXAD.
\end{abstract}

\section{Introduction}
Anomaly detection, which aims to identify data instances that do not conform to the expected behavior, is a classic machine learning task with numerous applications in various domains including fraud detection, intrusion detection, predictive maintenance, etc. Over the past decades, numerous methods have been proposed to tackle this challenging problem. Examples include one-class classification-based \cite{manevitz2001one,ruff2018deep}, nearest neighbor-based \cite{breunig2000lof}, clustering-based \cite{jiang2008clustering}, isolation-based \cite{liu2012isolation,hariri2019extended}, density-based \cite{liu2022unsupervised,feng2021time} and deep anomaly detection models based on autoencoders \cite{zhou2017anomaly,zong2018deep}, generative adversarial networks \cite{zenati2018adversarially,han2021gan}, to name a few. However, comparing to the vast body of literature on the detection task, anomaly explanation techniques have received relatively little attention so far~\cite{ruff_unifying_2021}. In fact, providing accurate explanations of why specific data instances are detected as anomalies is equally critical for many real-world applications. For instance, when an anomaly is reported by a fault detection application for a critical device in a factory, human operators need straightforward clues regarding the reported anomaly, and then can decide what next steps - such as fault diagnosis, predictive maintenance and system shutdown – should be taken. The required clues include which feature(s) is abnormal, why that feature(s) is abnormal. Furthermore, since model selection has a significant impact on the performance of anomaly detection tasks for tabular data \cite{han2022adbench}, providing model-agnostic anomaly explanation is considered more helpful than model-specific approaches that are only applicable to specific detection models.

To fill this gap, we propose a novel approach for efficient and accurate model-agnostic anomaly explanation for tabular data. Specifically, we leverage association rule mining \cite{agrawal1993mining} to learn Predicate-based Association Rules (PARs) which capture normal behaviors exhibited by training data. During inference, we efficiently find the precise PARs for explaining anomalies identified by arbitrary anomaly detection models. PARs provide intuitive explanations not only about which features of the anomaly instance are abnormal, but also why those features are abnormal. Our user study shows that the anomaly explanation form of PARs is better understood and favoured by regular anomaly detection system users compared with existing model-agnostic anomaly explanation options. In our experiments, we demonstrate that it is significantly more efficient to find PARs than anchors \cite{ribeiro2018anchors}, another rule-based explanation, for identified anomaly instances. Moreover, PARs are also far more accurate than anchors for anomaly explanation, meaning that they have considerably higher precision and recall when applied as anomaly detection rules on unseen data other than the anomaly instance on which they were originally derived for explanation. Additionally, we show that PARs can also achieve higher accuracy on abnormal feature identification compared with many state-of-the-art model-agnostic explanation methods including LIME \cite{ribeiro2016should}, SHAP \cite{lundberg2017unified}, COIN \cite{liu2018contextual} and ATON \cite{xu2021beyond}. We summarize the main contributions of this paper as follows:
\begin{itemize}
    \item We introduce PARs, a novel and informative explanation form for anomalies, and the form of PARs is more appealing to regular anomaly detection system users according to our user study.
    \item We provide a purely data-driven approach to efficiently constructing and finding the precise PARs for anomaly explanation.
    \item We reduce the expected time cost of deriving explanation rules for a single anomaly instance from \emph{tens of seconds} to \emph{less than one second} compared to anchors, which is critical for online anomaly detection and diagnosis applications.
    \item We conduct extensive experiments on public benchmark datasets to show that PARs can achieve more accurate anomaly explanation than existing model-agnostic explanation methods.
\end{itemize}

\section{PARs as Anomaly Explanations}
Given a black box anomaly detection model $f: \mathcal{X} \longrightarrow \mathcal{Y} $ where $\mathcal{Y} \in \{ 0,1\}$ with 0 indicating normality and 1 indicating abnormality, and an anomaly data instance $x \in \mathcal{X}$ with $f(x)=1$, let $\mathcal{F}$ be all the features of $\mathcal{X}$, our model-agnostic anomaly explanation aims to achieve two goals: 1) identify the abnormal feature subspace $\mathcal{F}_{sub} \in \mathcal{F}$ of $x$; 2) intuitively explain why the feature subspace $\mathcal{F}_{sub}$ of $x$ is abnormal. Both goals can be well achieved by PARs.

Specifically, a PAR is an association rule in the following form: $P \longrightarrow p$ where $P$ represents a set of antecedent predicates and $p$ represents a single consequent predicate such that $p \notin P$. PARs describe patterns of behavior that normal data instances should follow. An anomaly instance $x$ can be explained by a PAR if it violates the PAR, i.e., all the antecedent predicates of the PAR are satisfied but the consequent predicate is \textbf{not} satisfied by $x$. For example, in Table~\ref{tab:examples}, the PAR {\color{red}Level$>$10, Pump=ON $\longrightarrow$ Valve=Open} gives following explanation for the anomaly instance $x$=[11.1,ON,Close,25] with features $\mathcal{F}$=[Level, Pump, Valve, Temperature]: the Valve feature is abnormal because according to the PAR, when Level$>$10 and Pump=ON, then Valve should be Open, however, Valve=Close in $x$. This PAR actually describes a physical law that governs the behavior of a water tank system, where the outlet valve must be open to prevent the tank from exploding when the water level in the tank exceeds a certain threshold and the inlet pump remains on. Utilizing such rules with potential physical meanings for anomaly explanation is highly helpful for assisting users in understanding and diagnosing detected anomalies. Due to limited space, we put more examples of PARs for explaining anomaly instances in real-world usecases in the appendix. 

\begin{table*}
  \caption{Illustration of explanations of various model-agnostic methods to an anomaly instance ($x$): Level=11.1, Pump=ON, Valve=Close, Temperature=25 in a water tank condition monitoring dataset.}
  \label{tab:examples}
  \centering
  \begin{tabular}{ll}
    \toprule
    Method    & Explanation \\
    \midrule
    \multirow{2}{*}{SOAM} & {\color{red}\{Level, Pump, Valve\}}\\
    &   Level, Pump and Valve are abnormal features.\\
    \midrule
    COIN & {\color{red}Level: 0.3, Pump: 0.2, Valve: 0.4, Temperature: 0.1} \\
    ATON &  The abnormality weights for above features are [0.3, 0.2, 0.4, 0.1].\\
    \midrule
    \multirow{3}{*}{SHAP} & {\color{red}Level: 0.3$\uparrow$, Pump: 0.2$\uparrow$, Valve: 0.4$\uparrow$, Temperature: 0.1$\downarrow$} \\
    & Level, Pump and Valve features push the probability of prediction to anomaly higher \\
     & and the Temperature feature pushes the probability lower with different degrees. \\
    \midrule
    \multirow{3}{*}{LIME} & {\color{red}Level$>$10: 0.3$\uparrow$, Pump=ON: 0.2$\uparrow$, Valve=Close: 0.4$\uparrow$, Temperature$<$30: 0.1$\downarrow$}      \\
    & Level$>$10, Pump=ON and Valve=Close push the probability of prediction to anomaly\\
     &higher and Temperature$<$30 pushes the probability lower with different degrees. \\
     \midrule
    \multirow{2}{*}{Anchor}     & {\color{red}Level$>$10, Pump=ON, Valve=Close $\longrightarrow$ $f(x)=1$}     \\
    & If Level$>$10, Pump=ON and Valve=Close, then $x$ is classified as an anomaly. \\
    \midrule
    \multirow{3}{*}{PAR}     & {\color{red}Level$>$10, Pump=ON $\longrightarrow$ Valve=Open}         \\
    & If Level$>$10 and Pump=ON, then Valve should be OPEN. \\
    &If Level$>$10, Pump=ON but Valve$\neq$Open, then $x$ is classified as an anomaly. \\
    \bottomrule
  \end{tabular}
\end{table*}

\section{Related Work}
\subsection{Model-agnostic Anomaly Explanation}
The mainstream of prior art on model-agnostic anomaly explanation is to find the most anomalous/outlying feature subspace of anomaly instances in the score-and-search manner \cite{duan_mining_2015,vinh_discovering_2016,samariya2020new,samariya2020comprehensive}. Specifically, these methods identify abnormal features by searching for possible feature subspaces and compute the anomalous/outlying score of the anomaly instance in each subspace. However, since the number of possible subspaces increases exponentially with the growth of feature dimension, these methods are oftentimes too costly and potentially ineffective for high-dimensional data. Consequently, some methods are proposed to improve the efficiency and accuracy of abnormal feature identification. For example, SOAM \cite{ludtke2023outlying} is an outlying feature mining method which fits an Sum-Product Network (SPN) to model high-dimensional feature distributions, and leverages the tractability of marginal inference in SPNs to efficiently compute outlier scores in feature subsets. COIN \cite{liu2018contextual} transforms the anomaly explanation task into a classification problem which involves training a series of $l1-$norm classifiers that separate augmented anomalies from clusters of normal data in close proximity, and uses the classifiers' weights as anomaly contribution weights of features. ATON \cite{xu2021beyond} leverages the attention mechanism of neural networks to assign anomaly contribution weights to feature dimensions. Specifically, it utilizes a triplet deviation-based loss which estimates the separability of the anomaly instance and some heuristically sampled informative normal data within the triplets, and then a self-attention module is optimized to compute the contribution of each feature dimension to the anomaly instance. 

It is noteworthy to mention that most model-agnostic anomaly explanation methods like above focus on providing explanations for why the data instance is abnormal rather than why the anomaly detection algorithm has deemed it to be so. This is because the primary objective of anomaly explanation is to assist users for subsequent decision-making and diagnosis regarding the reported anomalies. PARs also focus on explaining the abnormality of data instead of the decision logic of the anomaly detection algorithm. 

\subsection{General Local Model-agnostic Explanation}
 Although not specifically designed for anomaly explanation tasks, some general local model-agnostic explanation methods that employ a perturbation-based strategy to generate local explanations for predictions of black box machine learning models, such as LIME \cite{ribeiro2016should}, SHAP \cite{lundberg2017unified} and Anchor \cite{ribeiro2018anchors}, are frequently used for anomaly explanation by considering anomaly detection as a binary classification problem. However, there are certain drawbacks of utilizing these general methods for anomaly explanation: 1) the perturbation-based strategy to generate local explanations for the algorithms is rather time-consuming, this makes such methods less applicable in typical online anomaly detection applications. 2) Determining the precise form to explain the decision logic of a complex anomaly detection algorithm is far more difficult than identifying the correct cause for the abnormality of data, which makes the derived explanations less robust than those methods directly explaining the abnormality of data.  

\subsection{Anomaly Explanation Forms}
It is important to emphasize that both informativeness and intuitiveness are crucial for anomaly explanation methods. To show the differences between the anomaly explanation forms provided by various model-agnostic methods, we illustrate the explanations of SOAM, COIN, ATON, SHAP, LIME, Anchor and PAR to an anomaly instance in a water tank condition monitoring dataset in Table~\ref{tab:examples}. As can be seen, most existing anomaly explanation methods only handle the abnormal feature identification task but do not provide intuitive explanations about why the corresponding features are abnormal. The only exception is Anchor, which also provides rule-based explanation about why a data instance is reported as anomaly. However, PARs are more informative than anchors. Referring to the example presented in Table~\ref{tab:examples}, when the predicate violation on the right-hand side is known, specifically Valve=Open, the user is able to identify the suspected abnormal feature. However, the anchor rule fails to accurately pinpoint the specific abnormal feature. Additionally, understanding the predicate violation on the right-hand side also informs the user about the correct behavior expected from the abnormal feature. In this instance, the valve should be in an open state rather than closed. This crucial information cannot be derived from the anchor rule when the variable can assume more than two potential values.

\subsection{Relation to Explainable Anomaly Detection}
There are existing methods which leverage association rules \cite{yairi2001fault,pal2017effectiveness,feng2019systematic} and other dependency-based methods \cite{paulheim2015decomposition,lu2020lopad,feng2020relsen} for explainable anomaly detection \cite{li2023survey}. However, we emphasize that our work is fundamentally different from those explainable anomaly detection methods. Specifically, those methods did not separate the anomaly detection task from the anomaly explanation task. Consequently, their application in practice is significantly limited due to their reliance on a relatively weak or less-general model for anomaly detection. In contrast, our approach designs a novel systematic framework for efficiently constructing and leveraging association rules exclusively for anomaly explanation. The decoupling of the anomaly explanation task from the anomaly detection task makes our method highly general and useful in practical applications.

\section{Learning PARs from Data}
In contrast to general local model-agnostic explanation methods where explanations are generated by a costly perturbation-based process during inference, we learn and store all PARs in the training stage. During inference, we simply need to efficiently find the precise PARs to explain the anomaly instance.

Learning PARs from a given dataset $\mathcal{D}$ mainly consists of two steps: \emph{predicate generation} and \emph{PAR mining}. Firstly, we derive a global predicate set $\mathcal{P}$ in the predicate generation step. We then transform each data instance $x \in \mathcal{D}$ as a set of satisfied predicates $P$ such that $P \subseteq \mathcal{P}$. Let $P_1,\ldots,P_{|\mathcal{D}|}$ represent the records of satisfied predicates for all data instances in $\mathcal{D}$, we then mine all PARs that satisfy a minimum support condition and a minimum confidence condition from the records. Specifically, let $P \longrightarrow p$ be a PAR, $\theta$ and $\gamma$ be the minimum support and minimum confidence thresholds respectively, we require $sup(P \longrightarrow p) > \theta$ and $conf(P \longrightarrow p) > \gamma$ where $sup(P \longrightarrow p)=\frac{\#(P \cup p)}{|\mathcal{D}|}$ measures the frequency of apparition of $P \cup p$ within the dataset, $conf(P \longrightarrow p)=\frac{sup(P \longrightarrow p)}{sup(P)}$ measures the percentage of data satisfying all the predicates in $P$ that also satisfy the consequent predicate $p$. Intuitively, a higher support for a PAR indicates a higher coverage of data whereas a higher confidence indicates a higher precision for explaining anomalies.

\subsection{Predicate Generation}
The quality of the global predicate set $\mathcal{P}$ is critical in our method. Specifically, the generated predicates should have high likelihoods leading to PARs. Furthermore, the form of generated predicates should be as simple as possible to maximize the interpretability of PARs. With these points in mind, we propose two algorithms for generating predicates, one specifically for categorical features and the other for numeric features. Before presenting the algorithms, it is important to highlight that in order for any generated predicate $p$ to have a nonzero probability of contributing to a PAR, $sup(p) > \theta$ is required. It can be seen that if the support of a predicate $p$ is less than $\theta$, then $p$ can never contribute to any PAR due to the anti-monotone constraint \cite{ng1998exploratory}.

\subsubsection{Predicate Generation for Categorical Features}
The algorithm of generating predicates for categorical features is straightforward. Let $\{ 1,\ldots,U \}$ be the set of observed values for a categorical feature $F^c$ in $\mathcal{D}$, we generate candidate predicates $p: F^c = u$ for all $u \in \{ 1,\ldots,U \}$. If $sup(p) > \theta$, we add $p$ to $\mathcal{P}$. Otherwise, we add $p$ to a list $\mathcal{L}$ which stores candidate predicates whose supports are less than the threshold. Then, we traverse all predicates in $\mathcal{L}$ and use the ``or''($``|"$) operator to generate combined predicates until their supports are larger than the threshold. For example, let $sup(p_1)<\theta$ and $sup(p_2)<\theta$, we generate a combined predicate $p:p_1 | p_2$ if $sup(p_1 | p_2) > \theta$. It is noteworthy to mention that we also prioritize the combination of predicates belonging to the same feature to optimize interpretability. Implementation details of the algorithm for categorical features is given in the appendix.

\subsubsection{Predicate Generation for Numeric Features}
To promote interpretability, we generate predicates for each numeric feature by a set of proposed cut-off values. For example, assuming there are three cut-off values $\tau_1$, $\tau_2$, $\tau_3$ for a numeric feature $F^n$ where $\tau_1 < \tau_2 < \tau_3$, we generate four predicates which are $p_1: F^n <\tau_1$, $p_2: \tau_1 \leq F^n <\tau_2$, $p_3: \tau_2 \leq F^n <\tau_3$ and $p_4: F^n \geq \tau_3$ and add them to $\mathcal{P}$ if the support for each predicate is larger than $\theta$. Moreover, we prefer cut-off values for a numeric feature which maximize the reduction of uncertainty in other features. In this way, predicates generated by such cut-off values could contribute to PARs with high likelihoods. Concretely, we employ a dependency-based approach which consists of following two steps to generate predicates for numeric features: 1) propose a set of candidate cut-off values by learning decision tree (DT) models on $\mathcal{D}$, 2) select cut-off values with higher impurity decrease to generate predicates.

\emph{Propose candidate cut-off values:} In this step, we learn two sets of DT models on $\mathcal{D}$ to propose candidate cut-off values. Specifically, Let $\mathcal{F}^c$ and $\mathcal{F}^n$ be the set of all categorical features and the set of all numeric features of the dataset, we first learn a DT classification model $DT(\mathcal{F}^n) \longrightarrow F^c$ for each categorical feature $F^c \in \mathcal{F}^c$ where all the numeric features $\mathcal{F}^n$ are used as input variables and $F^c$ is the prediction target. As for the second set, we learn a DT regression model $DT(\mathcal{F}^n_{-i}) \longrightarrow F^n_i$ for each numeric feature $F^n_i \in \mathcal{F}^n$ where all the remaining numeric features $\mathcal{F}^n_{-i}$ are used as input variables. Information gain and variance reduction \footnote{For DT regression models, the target feature is standardized before training such that variance reduction for different features are comparable.} are used to measure the quality of a split for the classification models and the regression models, respectively. We set the minimum number of samples required at a leaf node to be greater than $|\mathcal{D}|\times \theta$ as the stop criterion for the training of DT models. This avoids creating cut-off values which would definitely lead to predicates with support lower than $\theta$. Since each internal node in a learned DT model defines a split rule $\mathbf{1}_{(\tau,\infty)}(x_{F^n})$ where $\tau$ is a cut-off value on feature $F^n$ that directs an instance $x$ to the left/right child node, we can extract a tuple $(F^n,\tau,q_{\tau})$ from an internal node where $q_{\tau}$ is the impurity decrease of the node. Specifically,
\begin{eqnarray*}
    q_{\tau} = \frac{N}{|\mathcal{D}|} ( H - \frac{N_{left}}{N} H_{left} - \frac{N_{right}}{N} H_{right} )
\end{eqnarray*}where $N$, $N_{left}$ and $N_{right}$ are the number of data instances reaching the node, its left child and its right child; $H$, $H_{left}$ and $H_{right}$ are the impurity of the target feature for data reaching the node, its left child and right child, respectively. More specifically, impurity at a node is calculated as the entropy of the target feature for the data reaching the node for DT classification models and variance of the target feature for the data reaching the node for DT regression models. 


\emph{Select cut-off values for predicate generation:} It is beneficial to select cut-off values with higher $q_{\tau}$ values to generate predicates because such cut-off values reduce more uncertainty for the corresponding target feature and thus are more likely to contribute to PARs containing that feature. Therefore, after extracting all the cut-off values for a numeric feature $F^n$ from all the trained DT models, we arrange $F^n:(\tau_1, \dots,\tau_J)$ for the numeric feature such that $q_{\tau_1} \geq...\geq q_{\tau_J}$, i.e., the $q_{\tau}$ value for its cut-off values are sorted in a descending order. Then, we sequentially traverse the list of cut-off values and only keep a cut-off value for predicate generation for $F^n$ if its inclusion will not cancel the predicate of a cut-off value with a higher $q_{\tau}$ value. The logic of whether to keep a cut-off value for predicate generation is illustrated in Figure~\ref{fig:cutoff}. Implementation details of the whole algorithm is given in the appendix.

\begin{figure}
 \centering
 \includegraphics[width=.9\linewidth]{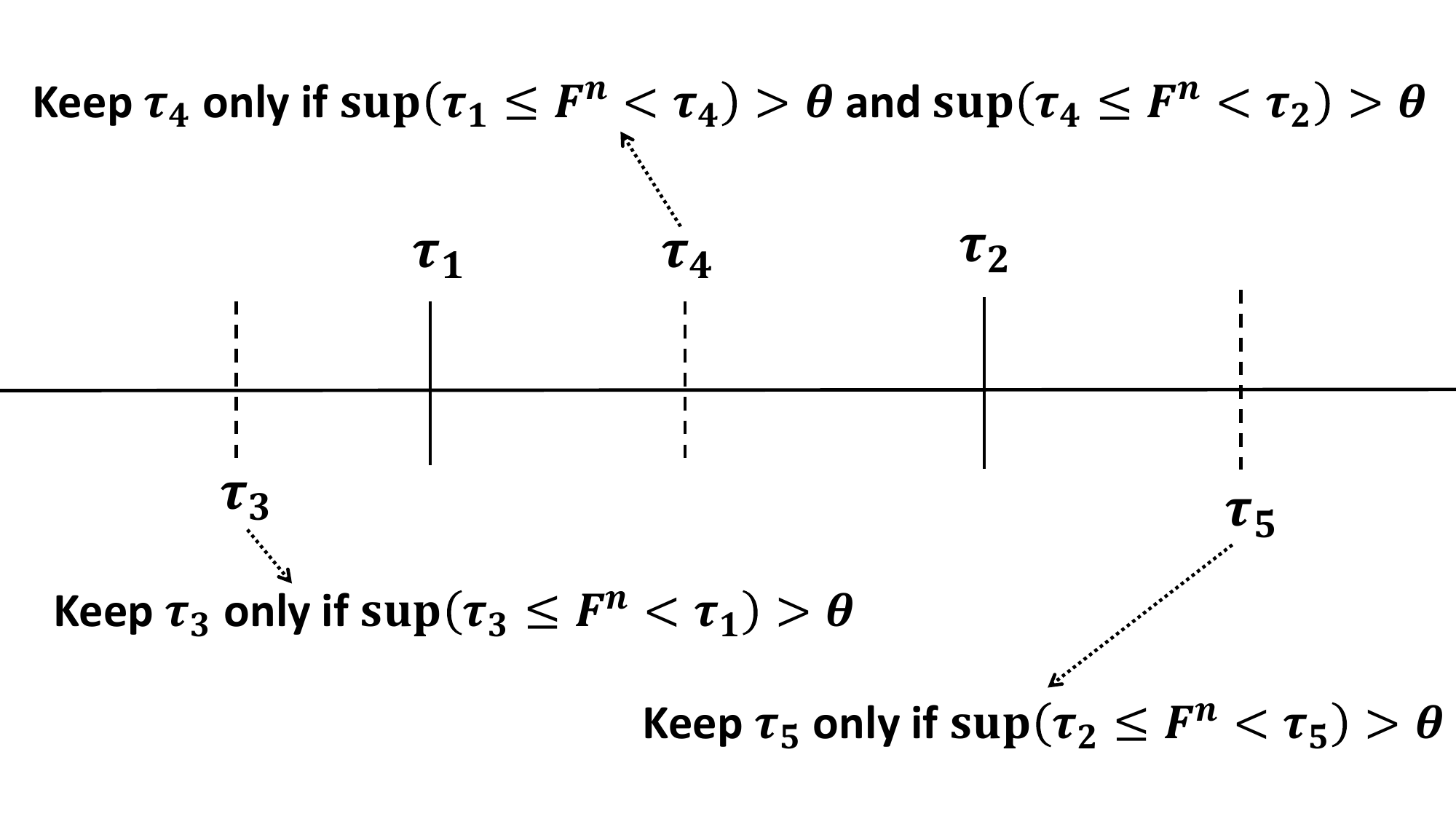}
 \caption{Illustration of the logic for whether keeping a cut-off value with a lower $q_{\tau}$ value for predicate generation. In the example, we assume $q_{\tau_1} \geq q_{\tau_2} \geq q_{\tau_3} \geq q_{\tau_4} \geq q_{\tau_5}$.}\label{fig:cutoff}
\end{figure}

\subsection{PAR Mining}
After obtaining the global predicate set $\mathcal{P}$, we transform each data instance $x \in \mathcal{D}$ as a set of satisfied predicates $P$ such that $P \subseteq \mathcal{P}$. Given the records of satisfied predicates for the data as $P_1,\ldots,P_{|\mathcal{D}|}$, mining PARs becomes an association rule mining problem. Concretely, we first find all frequent itemsets (predicate sets) with the minimum support threshold $\theta$ from the records using the FPGrowth algorithm \cite{han2000mining}. Then, for an arbitrary frequent predicate set $P$, regarding each $p \in P$, we partition $P$ into two parts $p$ and $P-p$, a PAR $P-p \longrightarrow p$ is generated if its confidence is larger than $\gamma$.

Besides, we also generate a set of special PARs called univariate PARs for explaining anomalies with simply out-of-range feature values. Concretely, for each categorical feature $F^c$, we generate an univariate PAR: $\emptyset \longrightarrow F^c \in \{1,\ldots,U\}$, where $\{1,\ldots,U\}$ is the set of seen values for $F^c$ in $\mathcal{D}$. For each numeric feature $F^n$, we generate an univariate PAR: $\emptyset \longrightarrow \mu-3\sigma \leq F^n \leq \mu+3\sigma$, where $\mu$ and $\sigma$ are the mean and standard deviation of the values for $F^n$ in $\mathcal{D}$.

\section{Finding Precise PARs for Explanation}
Let $\mathcal{A}$ be the set of all PARs learned from the training data $\mathcal{D}$. In the inference stage, when an anomaly instance $x$ is identified by a black box anomaly detection model $f$, we find the top-k PARs with the highest supports and confidences that are violated by $x$ to precisely explain the anomalous behavior of $x$. Concretely, the top-k PARs are selected as follows:
\begin{small}
\begin{eqnarray*}
\{A_1, ..., A_k \} = \underset{A \in \mathcal{A} \text{ and } A(x)=1}{\rm \arg \mathrm{Topk} } \quad \frac{sup(A)-\theta}{1-\theta}+ \lambda \frac{conf(A)-\gamma}{1-\gamma}
\end{eqnarray*}\end{small}where we call ($ \frac{sup(A)-\theta}{1-\theta}+ \lambda \frac{conf(A)-\gamma}{1-\gamma}$) the \emph{accuracy score} of PAR $A$ for anomaly explanation, in which $\lambda$ denotes the relative importance weight for the confidence with respect to the support; $A(x)=1$ if and only if $A$ is violated by $x$. Intuitively, this means we prefer to search for violated PARs with the highest coverage of data (support) and highest precision (confidence) to accurately explain anomalies.

Importantly, finding such top-k PARs for an arbitrary anomaly instance $x$ can be done rather efficiently. We sort all the PARs in $\mathcal{A}$ by their accuracy scores in a descending order in advance. Then, for an arbitrary identified anomaly instance $x$, we only need to sequentially traverse the sorted PARs in $\mathcal{A}$ until k violated PARs are found. 

\section{Hyperparameters and Potential Issues}
As noticed, we allow only one predicate in the right-hand side of a PAR. We outline the reason for doing this as follows: suppose there are three rules, $A_1$: $P \longrightarrow p_1$, $A_2$: $P \longrightarrow p_2$ and $A_3$: $P \longrightarrow \{p_1,p_2\}$. Then, due to the anti-monotone constraint, we have $sup(A_1)=\frac{\#(P \cup p_1)}{|\mathcal{D}|} \geq \frac{\#(P \cup \{p_1,p_2\})}{|\mathcal{D}|}=sup(A_3)$ and $conf(A_1)= \frac{sup(A_1)}{sup(P)} \geq \frac{sup(A_3)}{sup(P)} = conf(A_3)$, likewise $sup(A_2) \geq sup(A_3)$ and $conf(A_2)\geq conf(A_3)$. As a result, according to the accuracy score defined in the previous section, if the data satisfy $P$ and violates $p_1$ and/or $p_2$, then we prefer to pick $A_1$ and/or $A_2$ for anomaly explanation as they must have higher accuracy scores than $A_3$. Therefore, \emph{it is redundant to generating PARs with more than one predicate in the right-hand.} Regarding the maximum number of predicates in the left-hand side of a PAR, we limit it to four to make derived PARs not over-complicated for users.

Regarding the minimum confidence threshold $\gamma$ and the minimum support threshold $\theta$ in the PAR mining step, they define the lower limit for the confidence and support required for a PAR to be eligible for selection during the anomaly explanation process. Our algorithm is capable of identifying PARs with highest support and confidence using the accuracy score as the criterion for PAR selection. As a result, we do not recommend setting 
 $\gamma$ and $\theta$ to values that are too high because it will increase the risk of failing to find any PARs for explaining anomalies. In fact, we recommend to set $\gamma$ and $\theta$ to values less than which the found PARs are deemed to be entirely useless. Regarding $\lambda$, the importance weight for confidence, we recommend to set it to a value larger than one. This is because we generally prioritize high precision over high coverage of data for explaining anomalies, thus the confidence of PARs is much more important than their support in terms of facilitating accurate anomaly explanation. Throughout our experimentation, we fixed these hyperparameters to reasonable values $\theta= \max(10 / |\mathcal{D}|, 0.01)$, $\gamma=0.9$, $\lambda=5$, and achieved robust results. Note that by setting $\lambda$ to 5, we assign nearly identical accuracy score to a PAR with 100\% confidence and minimum support and another PAR with 98\% confidence and 100\% support. This weighting reflects our preference for confidence over support as the primary factor in selecting PARs. We leave an analysis of the sensitivity of our approach to these hyperparameters for future works.




\section{User Study}

We conducted a user study with 30 participants who use anomaly detection systems regularly in their work for different applications including equipment predictive maintenance, process condition monitoring and network intrusion detection. The sole objective of the user study is to investigate which anomaly explanation form is more useful to users under the assumption that all the explanations are accurate. Thus, we presented the users with various forms of anomaly explanations as shown in Table~\ref{tab:examples}, and asked them two questions: 1) ``Please rank the usefulness of different explanation forms in the table", and 2) ``Please provide reasons why you rank the particular explanation form as the best one". In the end, PAR was selected as the most useful form by 24 users for mainly two reasons: 1) rule-based format of PARs is more intuitive and understandable than other options, 2) PARs provide concrete information about the suspected abnormal feature. The average rankings for all algorithms are presented in Table~\ref{tab:user_study}, which shows that users generally prefer rule-based anomaly explanations, such as PAR and Anchor, over other alternatives.

\begin{table}
\begin{small}
  \caption{The average rankings of the usefulness of explanation forms provided by different methods in the user study.}
  \label{tab:user_study}
  \centering
  \begin{tabular}{lcccccc}
    \toprule
    Method    & SOAM & COIN & SHAP & LIME & Anchor & PAR \\
    \midrule
    Rank & 5.3 & 4.1 & 4.1 & 3.8 & 2.3 & \textbf{1.4} \\
    \bottomrule
  \end{tabular}
  \label{tab:explanation}
\end{small}
\end{table}

\begin{table}
\centering
\begin{small}
  \caption{Details of benchmark datasets}
  \label{tab:datasets}
  \centering
  \begin{tabular}{lcc}
    \toprule
    Dataset & \# Features & \# Samples   \\
    \midrule
    breastw & 9 & 683  \\
    cardio & 21 & 1831  \\
    Cardiotocography & 21 & 2114  \\ 
    fault & 27 & 1941  \\
    Ionosphere & 32 & 351 \\
    Lymphography & 18 & 148 \\
    magic & 10 & 19020 \\
    Pima & 8 & 768 \\
    satellite & 36 & 6435  \\
    satimage-2 & 36 & 5803 \\
    shuttle & 9 & 49097 \\
    skin & 3 & 245057  \\
    Stamps & 9 & 340  \\
    thyroid & 6 & 3772  \\
    WBC & 9 & 223  \\
    WDBC & 30 & 367  \\
    wine & 13 & 129  \\
    \bottomrule
  \end{tabular}
  \end{small}
\end{table}

\section{Experiments}
 In our experiments, two popular algorithms, Isolation Forecast (IF) \cite{liu2012isolation} and Autoencoder (AE) \cite{sakurada2014anomaly}, are selected as our representative anomaly detection models. The ADBench \cite{han2022adbench} is used as our benchmark datasets. Concretely, each tabular dataset in ADBench is split into a training set and a test set at a 4:1 ratio. We then train IF and AE models on the training set and tune anomaly thresholds for both models to achieve the best F$_1$ score on the testing set. We select all datasets on which both models can achieve at least 0.5 F$_1$ score on the testing set as our benchmark datasets for further experiments. This leads to 17 datasets being selected. The details of the select benchmark datasets are given in Table~\ref{tab:datasets}.



\subsection{Efficiency of Finding Explanation Rules}
We first investigate how efficient to find PARs for anomaly explanation. Anchor, which also provides rule-based anomaly explanation, is used as our baseline for comparison. We report the average time cost for computing the top 5 PARs and the anchor at each anomaly instance identified by IF and AE on the benchmark datasets in Table~\ref{tab:rule_comp}. Note that we only report the average metric across all the benchmark datasets in the table, detailed dataset-wise metrics are given in the appendix, and this also applies for all experiment results hereafter. As can be seen from the table, the average time cost of finding PARs is less than one second, whereas the average time cost for finding anchors is more than 25 seconds. This high computing efficiency makes PARs far more suitable for online anomaly explanation compared with Anchors.

\begin{table}
  \caption{The average time cost to compute the anchors and top 5 PARs at each anomaly instance in the benchmark datasets identified by IF and AE models, and the average precision, recall and F$_1$ score of using anchors and top 5 PARs computed at each anomaly instance to do anomaly detection in unseen data.}
  \label{tab:rule_comp}
  \centering
  \begin{tabular}{lccc}
    \toprule
    & Anchor & PAR& Improvement\\
    \midrule
    Avg. time cost (secs) & 25.71 & 0.22 & -99\% \\
    Avg. precision & 0.55 &  0.67 & +22\% \\ 
    Avg. recall & 0.21 &  0.39 & +86\% \\
    Avg. F$_1$ score & 0.23 &  0.42 & +83\% \\
    
    \bottomrule
  \end{tabular}
\end{table}

\subsection{Accuracy of Explanation Rules}
We then compare the accuracy of PARs to anchors. According to \cite{molnar2020interpretable}, the accuracy of an explanation is defined as ``How well does an explanation predict unseen data?". Thus for each dataset, to evaluate the accuracy of PARs and anchors, we use the top 5 PARs and the anchor computed for explaining a single anomaly instance as a anomaly detection model to detect anomalies in the remaining part of test set. We report the average performance, in terms of precision, recall and F$_1$ score, of using anchors and top 5 PARs computed at each identified anomaly instance to detect anomalies in unseen data in Table~\ref{tab:rule_comp}. As shown in the table, the selected PARs are much more accurate than anchors for anomaly explanation since on average PARs achieve 22\% improvement on precision, 86\% improvement on recall and 83\% improvement on F$_1$ score compared with anchors.

\subsection{Accuracy of Abnormal Feature Identification}

\begin{table}
  \caption{Accuracy of abnormal feature identification in terms of HitRate@100\% and HitRate@150\%.}
  \label{tab:afi_comp}
  \centering
  \begin{tabular}{lcccc}
    \toprule
    & \multicolumn{2}{c}{HitRate@100\%} & \multicolumn{2}{c}{HitRate@150\%} \\
    \cmidrule(r){2-3} \cmidrule(r){4-5}
    & Avg. & Avg. Rank & Avg. & Avg. Rank \\ 
    \midrule
    LIME & 0.39 & 4.47 & 0.48 & 4.35 \\
    SHAP & 0.59 & 2.24 & 0.68 & \textbf{2.12} \\
    COIN & 0.54 & 3.53 & 0.65 & 3.18 \\
    ATON & 0.62 & 2.53 & \textbf{0.71} & 2.47 \\
    PAR & \textbf{0.66} & \textbf{2.00} & 0.70 & 2.53 \\
    \bottomrule
  \end{tabular}
\end{table}

In this subsection, we study PARs' accuracy of abnormal feature identification compared with other state-of-the-art model-agnostic abnormal feature identification methods including LIME, SHAP, COIN and ATON. Specifically, for each identified anomaly, we output the features in the consequent predicates of the top 5 PARs as the suspected abnormal feature list. Since LIME, SHAP, COIN and ATON can output a feature weight vector per identified anomaly indicating the suspected anomaly contribution, we output a list of suspected abnormal features sorted by their anomaly contribution weights in a descending order for these methods.

\paragraph{Datasets preparation}Evaluating the accuracy of abnormal feature identification requires benchmark datasets with ground-truth annotations of abnormal feature subspace. To the best of our knowledge, there is no publicly available real-world tabular dataset with such annotations. As a result, we propose to use the 17 selected benchmark datasets for experiments. Specifically, we create ground-truth annotations of abnormal features for the benchmark datasets by randomly perturbing 1 to 3 features for the normal instances in the testing set. The perturbed features at each instance are labeled as abnormal features. We only apply different methods to identify abnormal features for perturbed data which are actually identified as anomalies by IF and AE models in our experiments.

\paragraph{Evaluation metrics and results}Similar to \cite{su2019robust}, we borrow the idea of HitRate@K for recommender systems \cite{yang2012top} as the metrics to evaluate the accuracy of abnormal feature identification. Specifically, we define a metric HitRate@P\%$=\frac{Hit@\floor{P\% \times |GF|}}{|GF|}$ where $|GF|$ is the size of ground-truth abnormal features and P can be 100 or 150. HitRate@P\% measures the number
of overlapping features between ground-truth abnormal features and the top $\floor{P\% \times |GF|}$ suspected abnormal features suggested by the anomaly explanation methods. For example, if ground-truth abnormal features are $\{2,6\}$ and the suspected abnormal feature list is [2, 3, 6, 1, 5, 4], the result is 0.5 for HitRate@100\% and 1.0 for HitRate@150\%. In Table~\ref{tab:afi_comp}, we report the metrics HitRate@100\% and HitRate@150\% for LIME, SHAP, COIN, ATON and PAR on the benchmark datasets. As can be seen from the table, PAR achieves the highest average HitRate@100\% and the second highest HitRate@150\% which demonstrate its high accuracy on abnormal feature identification. 


\begin{figure}
\begin{subfigure}[h]{0.49\linewidth}
\includegraphics[width=\linewidth]{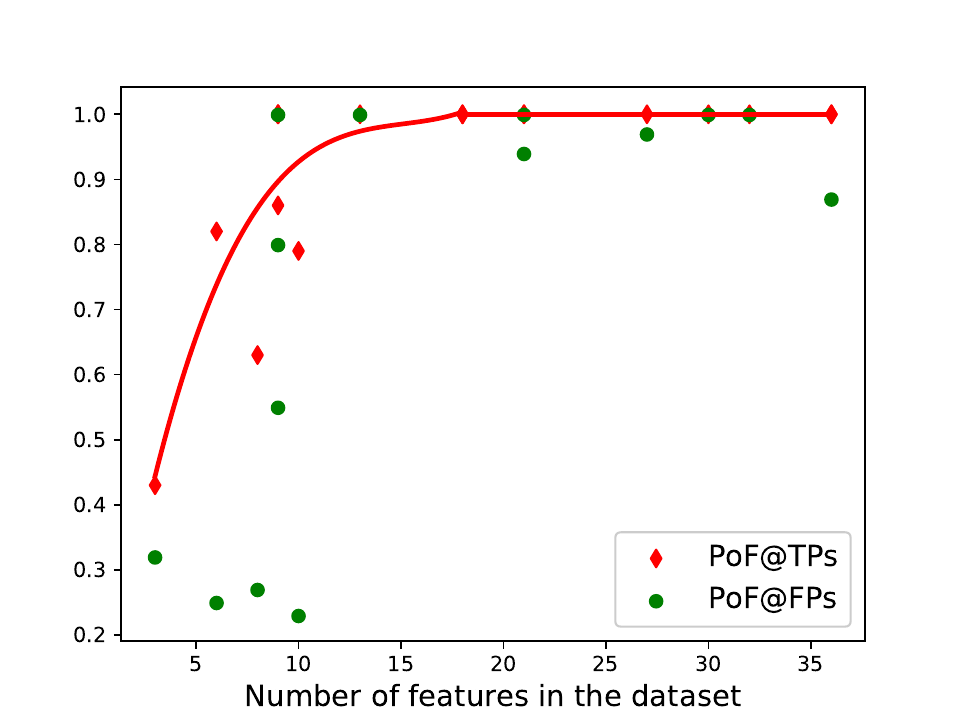}
\caption{}
\label{fig:pof}
\end{subfigure}
\hfill
\begin{subfigure}[h]{0.49\linewidth}
\includegraphics[width=\linewidth]{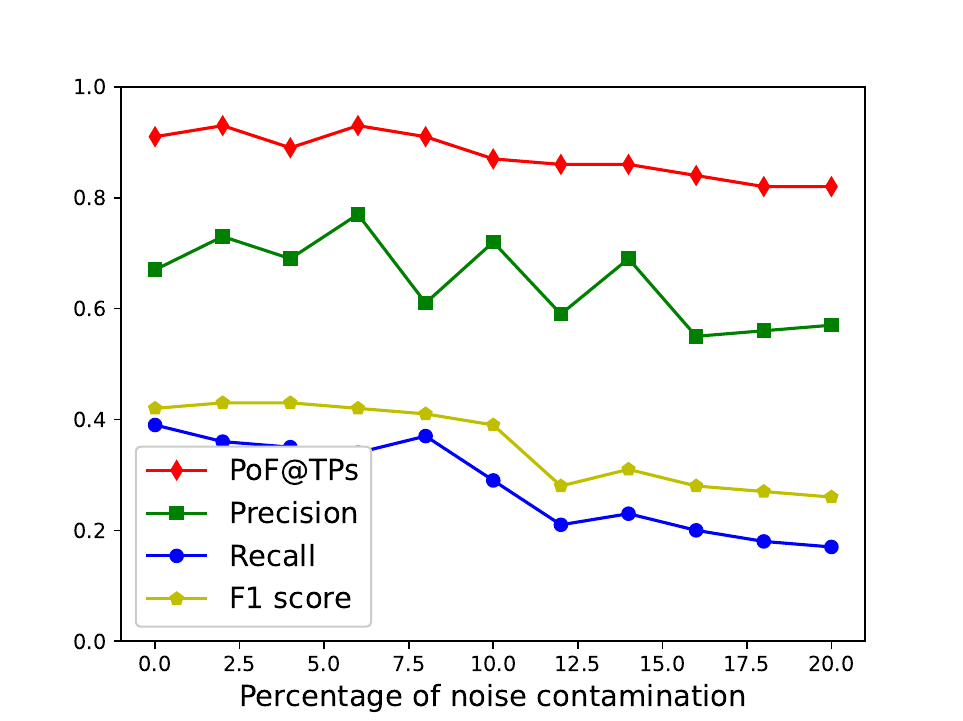}
\caption{}
\label{fig:noise}
\end{subfigure}%
\caption{(a): PoF@TPs and PoF@FPs for datasets with different number of features. (b): PoF@TPs and average accuracy of PARs with different percentage of noise contamination in the training data.}
\end{figure}

\subsection{Probability of Finding PARs for Anomalies}
In this subsection, we study the probability of finding at least one PAR (PoF for abbreviation) for explaining identified anomalies using our method. Concretely, we report PoF@TPs (PoF for True Positives) and PoF@FPs (PoF for True Positives) for predicated anomalies by IF and AE models on the benchmark datasets in Figure~\ref{fig:pof}. From the figure, we make two important observations: 1) PoF is discernibly higher when the predicted anomalies are TPs than when the predicted anomalies are FPs. This is desirable since we prefer to find PARs for explaining TPs. With respect to FPs, failure to finding a PAR for them actually gives the chance to mitigate their negative impact. 2) PoF is closely related to the feature dimension of datasets. This is as expected because we rely on the dependency between different features to construct PARs, thus the difficulty in finding PARs grows with fewer feature dimension. However, the good news is that we find PoF@TPs in 12 out of 17 datasets achieves 100\%. Moreover, there is a clear trend that when the feature dimension of dataset is larger than 10, PoF@TPs quickly converges to 100\% as shown in Figure~\ref{fig:pof}. 

\subsection{Impact of Noise Contamination}
To evaluate the robustness of our approach with respect to noise/anomaly contamination in the training data, we measure PoF@TPs and the accuracy of selected PARs when setting noise proportion within training data at levels ranging from $0\%$ to $20\%$. Specifically, we report PoF@TPs, the average precision, recall and F$_1$ score of applying top-5 PARs for explaining a single detected anomaly instance to anomaly detection on unseen data in Figure~\ref{fig:noise}. As can be seen, the impact of noise contamination on both PoF@TPs and the accuracy of selected PARs for anomaly explanation is limited when the noise contamination is less than $10\%$, implying that PARs are relatively robust to noise contamination in real-world anomaly detection scenarios. However, when the proportion of noise contamination exceeds $10\%$, which is an uncommon occurrence in real-world applications, the recall of selected PARs is primarily affected by the noise. This is likely due to the fact that the noise level has surpassed a certain threshold, causing a number of PARs to be unable to meet the minimum confidence threshold. This hypothesis is consistent with the decreasing trend of PoF@TPs as the proportion of noise contamination increases beyond $10\%$. Nevertheless, it is important to note that the precision of PARs still remains at a relatively high level, indicating that selected PARs remain rather reliable as they do not generate an excessive amount of false positives compared to PARs learned from noise-free data.

\section{Conclusion}

We introduced Predicate-based Association Rules (PARs), a novel and intuitive form of model-agnostic anomaly explanation. PARs not only highlight the suspected abnormal features, but also the reasons behind their abnormality. Our user study shows that PARs are better understood and preferred by regular anomaly detection system users compared with existing model-agnostic explanation options. We demonstrated the efficiency and the accuracy of PARs for anomaly explanation on various benchmark datasets. As people have increasingly highlighted the importance and imperativeness on providing tangible explanations in the anomaly detection field \cite{ruff_unifying_2021,pang2021toward}, PARs can make a highly valuable practical contribution to this domain.

\bibliography{aaai24}

\begin{thebibliography}{52}
\providecommand{\natexlab}[1]{#1}

\bibitem[{Agrawal, Imieli{\'n}ski, and Swami(1993)}]{agrawal1993mining}
Agrawal, R.; Imieli{\'n}ski, T.; and Swami, A. 1993.
\newblock Mining association rules between sets of items in large databases.
\newblock In \emph{Proceedings of the 1993 ACM SIGMOD international conference on Management of data}, 207--216.

\bibitem[{Agrawal, Srikant et~al.(1994)}]{agrawal1994fast}
Agrawal, R.; Srikant, R.; et~al. 1994.
\newblock Fast algorithms for mining association rules.
\newblock In \emph{Proc. 20th int. conf. very large data bases, VLDB}, volume 1215, 487--499. Citeseer.

\bibitem[{Barros, De~Carvalho, and Freitas(2015)}]{barros2015automatic}
Barros, R.~C.; De~Carvalho, A.~C.; and Freitas, A.~A. 2015.
\newblock \emph{Automatic design of decision-tree induction algorithms}.
\newblock Springer.

\bibitem[{Breiman et~al.(2017)Breiman, Friedman, Olshen, and Stone}]{breiman2017classification}
Breiman, L.; Friedman, J.~H.; Olshen, R.~A.; and Stone, C.~J. 2017.
\newblock \emph{Classification and regression trees}.
\newblock Routledge.

\bibitem[{Breunig et~al.(2000)Breunig, Kriegel, Ng, and Sander}]{breunig2000lof}
Breunig, M.~M.; Kriegel, H.-P.; Ng, R.~T.; and Sander, J. 2000.
\newblock LOF: identifying density-based local outliers.
\newblock In \emph{Proceedings of the 2000 ACM SIGMOD international conference on Management of data}, 93--104.

\bibitem[{Duan et~al.(2015)Duan, Tang, Pei, Bailey, Campbell, and Tang}]{duan_mining_2015}
Duan, L.; Tang, G.; Pei, J.; Bailey, J.; Campbell, A.; and Tang, C. 2015.
\newblock Mining outlying aspects on numeric data.
\newblock \emph{Data Mining and Knowledge Discovery}, 29(5): 1116--1151.

\bibitem[{Feng et~al.(2020)Feng, Liang, Schneegass, and Tian}]{feng2020relsen}
Feng, C.; Liang, X.; Schneegass, D.; and Tian, P. 2020.
\newblock RelSen: An Optimization-based Framework for Simultaneously Sensor Reliability Monitoring and Data Cleaning.
\newblock In \emph{Proceedings of the 29th ACM International Conference on Information \& Knowledge Management}, 345--354.

\bibitem[{Feng et~al.(2019)Feng, Palleti, Mathur, and Chana}]{feng2019systematic}
Feng, C.; Palleti, V.~R.; Mathur, A.; and Chana, D. 2019.
\newblock A Systematic Framework to Generate Invariants for Anomaly Detection in Industrial Control Systems.
\newblock In \emph{26th Annual Network and Distributed System Security Symposium, {NDSS} 2019, San Diego, California, USA, February 24-27, 2019}. The Internet Society.

\bibitem[{Feng and Tian(2021)}]{feng2021time}
Feng, C.; and Tian, P. 2021.
\newblock Time series anomaly detection for cyber-physical systems via neural system identification and bayesian filtering.
\newblock In \emph{Proceedings of the 27th ACM SIGKDD Conference on Knowledge Discovery \& Data Mining}, 2858--2867.

\bibitem[{Freitas(2014)}]{freitas2014comprehensible}
Freitas, A.~A. 2014.
\newblock Comprehensible classification models: a position paper.
\newblock \emph{ACM SIGKDD explorations newsletter}, 15(1): 1--10.

\bibitem[{Han, Pei, and Yin(2000)}]{han2000mining}
Han, J.; Pei, J.; and Yin, Y. 2000.
\newblock Mining frequent patterns without candidate generation.
\newblock \emph{ACM sigmod record}, 29(2): 1--12.

\bibitem[{Han et~al.(2022)Han, Hu, Huang, Jiang, and Zhao}]{han2022adbench}
Han, S.; Hu, X.; Huang, H.; Jiang, M.; and Zhao, Y. 2022.
\newblock Adbench: Anomaly detection benchmark.
\newblock \emph{Advances in Neural Information Processing Systems}, 35: 32142--32159.

\bibitem[{Han, Chen, and Liu(2021)}]{han2021gan}
Han, X.; Chen, X.; and Liu, L.-P. 2021.
\newblock Gan ensemble for anomaly detection.
\newblock In \emph{Proceedings of the AAAI Conference on Artificial Intelligence}, volume~35, 4090--4097.

\bibitem[{Hariri, Kind, and Brunner(2019)}]{hariri2019extended}
Hariri, S.; Kind, M.~C.; and Brunner, R.~J. 2019.
\newblock Extended isolation forest.
\newblock \emph{IEEE Transactions on Knowledge and Data Engineering}, 33(4): 1479--1489.

\bibitem[{Jiang and An(2008)}]{jiang2008clustering}
Jiang, S.-y.; and An, Q.-b. 2008.
\newblock Clustering-based outlier detection method.
\newblock In \emph{2008 Fifth international conference on fuzzy systems and knowledge discovery}, volume~2, 429--433. IEEE.

\bibitem[{Kass(1980)}]{kass1980exploratory}
Kass, G.~V. 1980.
\newblock An exploratory technique for investigating large quantities of categorical data.
\newblock \emph{Journal of the Royal Statistical Society: Series C (Applied Statistics)}, 29(2): 119--127.

\bibitem[{Laurent and Rivest(1976)}]{laurent1976constructing}
Laurent, H.; and Rivest, R.~L. 1976.
\newblock Constructing optimal binary decision trees is NP-complete.
\newblock \emph{Information processing letters}, 5(1): 15--17.

\bibitem[{Li, Zhu, and Van~Leeuwen(2023)}]{li2023survey}
Li, Z.; Zhu, Y.; and Van~Leeuwen, M. 2023.
\newblock A survey on explainable anomaly detection.
\newblock \emph{ACM Transactions on Knowledge Discovery from Data}, 18(1): 1--54.

\bibitem[{Liu, Tan, and Zhou(2022)}]{liu2022unsupervised}
Liu, B.; Tan, P.-N.; and Zhou, J. 2022.
\newblock Unsupervised Anomaly Detection by Robust Density Estimation.
\newblock In \emph{Proceedings of the AAAI Conference on Artificial Intelligence}.

\bibitem[{Liu, Ting, and Zhou(2012)}]{liu2012isolation}
Liu, F.~T.; Ting, K.~M.; and Zhou, Z.-H. 2012.
\newblock Isolation-based anomaly detection.
\newblock \emph{ACM Transactions on Knowledge Discovery from Data (TKDD)}, 6(1): 1--39.

\bibitem[{Liu, Shin, and Hu(2018)}]{liu2018contextual}
Liu, N.; Shin, D.; and Hu, X. 2018.
\newblock Contextual outlier interpretation.
\newblock In \emph{Proceedings of the 27th International Joint Conference on Artificial Intelligence}, 2461--2467.

\bibitem[{Lu et~al.(2020)Lu, Liu, Li, Le, and Liu}]{lu2020lopad}
Lu, S.; Liu, L.; Li, J.; Le, T.~D.; and Liu, J. 2020.
\newblock Lopad: A local prediction approach to anomaly detection.
\newblock In \emph{Pacific-Asia Conference on Knowledge Discovery and Data Mining}, 660--673. Springer.

\bibitem[{L{\"u}dtke, Bartelt, and Stuckenschmidt(2023)}]{ludtke2023outlying}
L{\"u}dtke, S.; Bartelt, C.; and Stuckenschmidt, H. 2023.
\newblock Outlying Aspect Mining via Sum-Product Networks.
\newblock In \emph{Pacific-Asia Conference on Knowledge Discovery and Data Mining}, 27--38. Springer.

\bibitem[{Lundberg and Lee(2017)}]{lundberg2017unified}
Lundberg, S.~M.; and Lee, S.-I. 2017.
\newblock A unified approach to interpreting model predictions.
\newblock \emph{Advances in neural information processing systems}, 30.

\bibitem[{Manevitz and Yousef(2001)}]{manevitz2001one}
Manevitz, L.~M.; and Yousef, M. 2001.
\newblock One-class SVMs for document classification.
\newblock \emph{Journal of machine Learning research}, 2(Dec): 139--154.

\bibitem[{Mathur and Tippenhauer(2016)}]{mathur2016swat}
Mathur, A.~P.; and Tippenhauer, N.~O. 2016.
\newblock SWaT: A water treatment testbed for research and training on ICS security.
\newblock In \emph{2016 international workshop on cyber-physical systems for smart water networks (CySWater)}, 31--36. IEEE.

\bibitem[{Molnar(2020)}]{molnar2020interpretable}
Molnar, C. 2020.
\newblock \emph{Interpretable machine learning}.
\newblock Lulu. com.

\bibitem[{Nanfack, Temple, and Fr{\'e}nay(2022)}]{nanfack2022constraint}
Nanfack, G.; Temple, P.; and Fr{\'e}nay, B. 2022.
\newblock Constraint Enforcement on Decision Trees: a Survey.
\newblock \emph{ACM Computing Surveys (CSUR)}.

\bibitem[{Ng et~al.(1998)Ng, Lakshmanan, Han, and Pang}]{ng1998exploratory}
Ng, R.~T.; Lakshmanan, L.~V.; Han, J.; and Pang, A. 1998.
\newblock Exploratory mining and pruning optimizations of constrained associations rules.
\newblock \emph{ACM Sigmod Record}, 27(2): 13--24.

\bibitem[{Nunes et~al.(2020)Nunes, De~Craene, Langet, Camara, and Jonsson}]{nunes2020learning}
Nunes, C.; De~Craene, M.; Langet, H.; Camara, O.; and Jonsson, A. 2020.
\newblock Learning decision trees through Monte Carlo tree search: An empirical evaluation.
\newblock \emph{Wiley Interdisciplinary Reviews: Data Mining and Knowledge Discovery}, 10(3): e1348.

\bibitem[{Pal, Adepu, and Goh(2017)}]{pal2017effectiveness}
Pal, K.; Adepu, S.; and Goh, J. 2017.
\newblock Effectiveness of association rules mining for invariants generation in cyber-physical systems.
\newblock In \emph{2017 IEEE 18th International Symposium on High Assurance Systems Engineering (HASE)}, 124--127. IEEE.

\bibitem[{Pang and Aggarwal(2021)}]{pang2021toward}
Pang, G.; and Aggarwal, C. 2021.
\newblock Toward explainable deep anomaly detection.
\newblock In \emph{Proceedings of the 27th ACM SIGKDD Conference on Knowledge Discovery \& Data Mining}, 4056--4057.

\bibitem[{Paulheim and Meusel(2015)}]{paulheim2015decomposition}
Paulheim, H.; and Meusel, R. 2015.
\newblock A decomposition of the outlier detection problem into a set of supervised learning problems.
\newblock \emph{Machine Learning}, 100(2): 509--531.

\bibitem[{Pedregosa et~al.(2011)Pedregosa, Varoquaux, Gramfort, Michel, Thirion, Grisel, Blondel, Prettenhofer, Weiss, Dubourg et~al.}]{pedregosa2011scikit}
Pedregosa, F.; Varoquaux, G.; Gramfort, A.; Michel, V.; Thirion, B.; Grisel, O.; Blondel, M.; Prettenhofer, P.; Weiss, R.; Dubourg, V.; et~al. 2011.
\newblock Scikit-learn: Machine learning in Python.
\newblock \emph{the Journal of machine Learning research}, 12: 2825--2830.

\bibitem[{Quinlan(1986)}]{quinlan1986induction}
Quinlan, J.~R. 1986.
\newblock Induction of decision trees.
\newblock \emph{Machine learning}, 1(1): 81--106.

\bibitem[{Quinlan(2014)}]{quinlan2014c4}
Quinlan, J.~R. 2014.
\newblock \emph{C4. 5: programs for machine learning}.
\newblock Elsevier.

\bibitem[{Ribeiro, Singh, and Guestrin(2016)}]{ribeiro2016should}
Ribeiro, M.~T.; Singh, S.; and Guestrin, C. 2016.
\newblock " Why should i trust you?" Explaining the predictions of any classifier.
\newblock In \emph{Proceedings of the 22nd ACM SIGKDD international conference on knowledge discovery and data mining}, 1135--1144.

\bibitem[{Ribeiro, Singh, and Guestrin(2018)}]{ribeiro2018anchors}
Ribeiro, M.~T.; Singh, S.; and Guestrin, C. 2018.
\newblock Anchors: High-precision model-agnostic explanations.
\newblock In \emph{Proceedings of the AAAI conference on artificial intelligence}, volume~32.

\bibitem[{Ruff et~al.(2021)Ruff, Kauffmann, Vandermeulen, Montavon, Samek, Kloft, Dietterich, and Müller}]{ruff_unifying_2021}
Ruff, L.; Kauffmann, J.~R.; Vandermeulen, R.~A.; Montavon, G.; Samek, W.; Kloft, M.; Dietterich, T.~G.; and Müller, K.-R. 2021.
\newblock A {Unifying} {Review} of {Deep} and {Shallow} {Anomaly} {Detection}.
\newblock \emph{Proceedings of the IEEE}, 109(5): 756--795.

\bibitem[{Ruff et~al.(2018)Ruff, Vandermeulen, Goernitz, Deecke, Siddiqui, Binder, M{\"u}ller, and Kloft}]{ruff2018deep}
Ruff, L.; Vandermeulen, R.; Goernitz, N.; Deecke, L.; Siddiqui, S.~A.; Binder, A.; M{\"u}ller, E.; and Kloft, M. 2018.
\newblock Deep one-class classification.
\newblock In \emph{International conference on machine learning}, 4393--4402. PMLR.

\bibitem[{Sakurada and Yairi(2014)}]{sakurada2014anomaly}
Sakurada, M.; and Yairi, T. 2014.
\newblock Anomaly detection using autoencoders with nonlinear dimensionality reduction.
\newblock In \emph{Proceedings of the MLSDA 2014 2nd workshop on machine learning for sensory data analysis}, 4--11.

\bibitem[{Samariya et~al.(2020)Samariya, Aryal, Ting, and Ma}]{samariya2020new}
Samariya, D.; Aryal, S.; Ting, K.~M.; and Ma, J. 2020.
\newblock A new effective and efficient measure for outlying aspect mining.
\newblock In \emph{Web Information Systems Engineering--WISE 2020: 21st International Conference, Amsterdam, The Netherlands, October 20--24, 2020, Proceedings, Part II 21}, 463--474. Springer.

\bibitem[{Samariya, Ma, and Aryal(2020)}]{samariya2020comprehensive}
Samariya, D.; Ma, J.; and Aryal, S. 2020.
\newblock A comprehensive survey on outlying aspect mining methods.
\newblock \emph{arXiv preprint arXiv:2005.02637}.

\bibitem[{Sharma and Tivari(2012)}]{sharma2012survey}
Sharma, A.; and Tivari, N. 2012.
\newblock A survey of association rule mining using genetic algorithm.
\newblock \emph{International Journal of Computer Applications \& Information Technology}, 1(2): 5--11.

\bibitem[{Su et~al.(2019)Su, Zhao, Niu, Liu, Sun, and Pei}]{su2019robust}
Su, Y.; Zhao, Y.; Niu, C.; Liu, R.; Sun, W.; and Pei, D. 2019.
\newblock Robust anomaly detection for multivariate time series through stochastic recurrent neural network.
\newblock In \emph{Proceedings of the 25th ACM SIGKDD international conference on knowledge discovery \& data mining}, 2828--2837.

\bibitem[{Vinh et~al.(2016)Vinh, Chan, Romano, Bailey, Leckie, Ramamohanarao, and Pei}]{vinh_discovering_2016}
Vinh, N.~X.; Chan, J.; Romano, S.; Bailey, J.; Leckie, C.; Ramamohanarao, K.; and Pei, J. 2016.
\newblock Discovering outlying aspects in large datasets.
\newblock \emph{Data Mining and Knowledge Discovery}, 30(6): 1520--1555.

\bibitem[{Xu et~al.(2021)Xu, Wang, Jian, Huang, Wang, Liu, and Li}]{xu2021beyond}
Xu, H.; Wang, Y.; Jian, S.; Huang, Z.; Wang, Y.; Liu, N.; and Li, F. 2021.
\newblock Beyond outlier detection: Outlier interpretation by attention-guided triplet deviation network.
\newblock In \emph{Proceedings of the Web Conference 2021}, 1328--1339.

\bibitem[{Yairi, Kato, and Hori(2001)}]{yairi2001fault}
Yairi, T.; Kato, Y.; and Hori, K. 2001.
\newblock Fault detection by mining association rules from house-keeping data.
\newblock In \emph{proceedings of the 6th International Symposium on Artificial Intelligence, Robotics and Automation in Space}, volume~18, 21. Citeseer.

\bibitem[{Yang et~al.(2012)Yang, Steck, Guo, and Liu}]{yang2012top}
Yang, X.; Steck, H.; Guo, Y.; and Liu, Y. 2012.
\newblock On top-k recommendation using social networks.
\newblock In \emph{Proceedings of the sixth ACM conference on Recommender systems}, 67--74.

\bibitem[{Zenati et~al.(2018)Zenati, Romain, Foo, Lecouat, and Chandrasekhar}]{zenati2018adversarially}
Zenati, H.; Romain, M.; Foo, C.-S.; Lecouat, B.; and Chandrasekhar, V. 2018.
\newblock Adversarially learned anomaly detection.
\newblock In \emph{2018 IEEE International conference on data mining (ICDM)}, 727--736. IEEE.

\bibitem[{Zhou and Paffenroth(2017)}]{zhou2017anomaly}
Zhou, C.; and Paffenroth, R.~C. 2017.
\newblock Anomaly detection with robust deep autoencoders.
\newblock In \emph{Proceedings of the 23rd ACM SIGKDD international conference on knowledge discovery and data mining}, 665--674.

\bibitem[{Zong et~al.(2018)Zong, Song, Min, Cheng, Lumezanu, Cho, and Chen}]{zong2018deep}
Zong, B.; Song, Q.; Min, M.~R.; Cheng, W.; Lumezanu, C.; Cho, D.; and Chen, H. 2018.
\newblock Deep autoencoding gaussian mixture model for unsupervised anomaly detection.
\newblock In \emph{International conference on learning representations}.

\end{thebibliography}

\clearpage
\appendix

\section{Appendix}
\subsection{Association rule mining}
Association rule mining, one of the most important data mining techniques, is used to discover the frequently occurring patterns in the database \cite{agrawal1993mining}. The main aim of association rule mining is to find out the interesting relationships and correlations among the different items of the database. Specifically, let $I=\{i_1, i_2, \ldots i_m \}$ be a set of items and $D$ be a set of transactions, where each transaction $T$ is a set of items such that $T \subseteq I$. An association rule is expressed as $X \Rightarrow Y$ where $X,Y \subseteq I$ and $X \cap Y=\emptyset$. Furthermore, there are two basic measures for an association rule: support (s) and confidence (c). A rule has support $s$ if $s$ proportion of the transactions in $D$ contains $X \cup Y$. A rule $X \Rightarrow Y$ has confidence $c$, if $c$ proportion of transactions in $D$ that support X also support Y. Given a set of transactions $D$ (the database), the problem of mining association rules is to discover all association rules that have support and confidence greater than the user-specified minimum support (called \emph{minsup}) and minimum confidence (called \emph{minconf}). More specifically, association rules are commonly generated using the following two steps:
1) Find all the frequent itemsets whose support is larger than \emph{minsup}.
2) Based on these frequent itemsets, association rules which have confidence above \emph{minconf} are generated. The first step is much more difficult than the second step. Many algorithms have been developed to mine frequent itemsets, including Apriori \cite{agrawal1994fast}, FP-Growth \cite{han2000mining} and genetic algorithms \cite{sharma2012survey}, etc. 

\subsection{Decision tree learning}
Decision trees (DTs) are a family of machine learning algorithms primarily designed for classification and regression. Their representability and ability to produce rules with relevant attributes make them the most commonly used technique when seeking interpretable machine learning models \cite{freitas2014comprehensible,nanfack2022constraint}. 

Specifically, let $X$ be the input variables with $M$ dimensions $X_i$ where $i=1,\ldots,M$, Y be the output variable ($Y \in \{1,\dots,C\}$ for classification, $Y \in \mathcal{R}$ for regression), $\mathcal{D}$ be the dataset formed by sampling from the unknown joint distribution $P_{XY}$. A DT consists of a hierarchy of internal nodes with defined splitting rules based on $X$, and a set of leaf nodes with predictions about $Y$. Splitting rules can involve one variable each time leading to univariate DTs, or multiple variables each time leading to multivariate DTs. To promote interpretability, we only consider univariate DTs in this work. Following \cite{nunes2020learning}, we denote a decision rule on a single variable $X_i$ as $f(\mathbf{x}) = \mathbf{1}_A(x_i)$ where $\mathbf{1}_A(x_i)$ is an indicator function, taking value 1 for $x_i \in A$ and 0 otherwise. For numerical $X_i$, we have:
$f(\mathbf{x}) = \mathbf{1}_{(\tau,\infty)}(x_i)$
where $\tau$ is the selected cut-off value. Intuitively, a 0/1 outcome directs the instance $\mathbf{x}$ to the left/right child node.

The learning of DTs is mainly composed by induction and pruning. Induction is about learning the DT structure and its splitting rules. \cite{laurent1976constructing} shows that learning an optimal DT that maximizes prediction performance while minimizing the size of the tree is NP-complete owing to the discrete and sequential nature of the splits. As a result, standard DT algorithms such as CHAID \cite{kass1980exploratory}, CART \cite{breiman2017classification}, ID3 \cite{quinlan1986induction}, and C4.5 \cite{quinlan2014c4} learn a DT by following locally optimal induction strategies. Specifically, locally optimal induction selects splitting rules that maximize an objective at each node, e.g., variance reduction for regression, information gain for classification. The splitting procedure stops when a specific criterion, e.g., the maximum depth of the tree and the minimum number of samples required to be at a leaf node, is reached and a leaf node is created. By using this locally optimal search heuristic, learned greedy trees can be very accurate but significantly overfit. To avoid that, pruning techniques are commonly applied to find a trade-off between reducing the complexity of the tree and maintaining a certain level of accuracy \cite{barros2015automatic,nanfack2022constraint}.

\subsection{Examples of Using PARs for Anomaly Explanation}
To demonstrate the informativeness and intuitivenenss of PARs for anomaly explanation, we conducted additional experiments applying PARs to explain detected anomalies in SWAT, a dataset collected from a real-world water treatment testbed \cite{mathur2016swat}. Note that we did not use the same datasets from ADBench only because they do not have feature names with physical meanings, which makes them less useful for illustrating the informativeness of PARs. We present selected PARs (using the same hyperparameter values as in the paper) for explaining three example anomaly instances with labeled ground-truth abnormal features as below:
\begin{itemize}
    \item Anomaly instance 1:
    \begin{itemize}
        \item Ground-truth abnormal features: MV101
        \item Top-1 selected PAR for anomaly explanation: LIT101$\geq$811.18$\longrightarrow$MV101=1
        \item Translation: If the reading of level indicator LIT101 is larger than 811.18, then the state of valve MV101 should be 1, however, MV101 is not in state 1. This means that MV101 is suspected to be abnormal.
    \end{itemize}
    \item Anomaly instance 2:
    \begin{itemize}
        \item Ground-truth abnormal features: AIT202, P203
        \item Top-1 selected PAR for anomaly explanation: $\emptyset \longrightarrow$8.21$\leq$AIT202$\leq$8.84
        \item Translation: The reading of AIT202 should be in the range between 8.21 and 8.84, the current reading of AIT202 is not within the correct range. This means AIT202 is suspected to be abnormal.
        \item Top-2 selected PAR for anomaly explanation: P205=1$\longrightarrow$P203=1
        \item Translation: If pump P205 is in state 1, then pump P203 should also be in state 1. However, pump P203 is not in state 1. This means P203 is suspected to be abnormal.
    \end{itemize}
    \item Anomaly instance 3:
    \begin{itemize}
        \item Ground-truth abnormal features: LIT101
        \item Top-1 selected PAR for anomaly explanation: AIT202$<$8.50, MV101=2, 188.16$\leq$PIT503$\leq$189.43, 824.64$\leq$LIT301$\leq$940.62$\longrightarrow$LIT101$<$537.59
        \item Translation: If sensor reading AIT202 is less than 8.5, valve MV101 is in state 2, sensor reading PIT503 is between 188.16 and 189.43, sensor reading LIT301 is between 824.64 and 940.62, then sensor reading for LIT101 should be less than 537.59. However, LIT101 is larger than 537.59. This means LIT101 is suspected to be abnormal.
    \end{itemize}
\end{itemize}

\subsection{Implementation Details for Predicate Generation Algorithms}

Algorithm~\ref{alg:algorithmcate} gives the details of generating predicates for categorical features. Algorithm~\ref{alg:predicate_cont} gives the details of predicate generation for numeric features.

\begin{algorithm*}
  \caption{Predicate generation for categorical features}
  \label{alg:algorithmcate}
  \begin{algorithmic}[1]
  	\REQUIRE The dataset $\mathcal{D}$, the minimum support threshold $\theta$, the categorical feature set $\mathcal{F}^c$
  	\STATE $\mathcal{P}\gets \emptyset$, $\mathcal{L}\gets \emptyset$
  	\FOR{$i=1,\ldots,|\mathcal{F}^c|$}
            \STATE $\mathcal{T}\gets \emptyset$
  	    \STATE Let $\{1,\ldots,U\}$ be the set of possible values for a categorical feature $F^c_i$
  	    \FOR{$u=1,\ldots,U$}
          	\STATE Generate predicate $p$: $F^c_i=u$
          	\IF{$sup(p)>\theta$}
          	    \STATE Add $p$ to $\mathcal{P}$
          	\ELSE
          	    \STATE Add $p$ to $\mathcal{T}$
          	\ENDIF
        \ENDFOR
        \IF{$|\mathcal{T}|>1$}
            \STATE Generate predicate $p$: $p_1 |\dots| p_{|\mathcal{T}|}$ \COMMENT{combine predicates for a single feature}
            \IF{$sup(p)>\theta$}
                \STATE Add $p$ to $\mathcal{P}$
            \ELSE
                \STATE Add $p$ to $\mathcal{L}$
            \ENDIF
        \ELSIF{$|\mathcal{T}|=1$}
            \STATE Add $p$ in $\mathcal{T}$ to $\mathcal{L}$
        \ENDIF
    \ENDFOR
    
    \STATE $k \gets 1$
    \FOR{$j=2,\ldots,|\mathcal{L}|$}
        \IF{$sup(p_k |\dots| p_j) > \theta$}
            \IF{$sup(p_{j+1} |\dots| p_{|\mathcal{L}|}) > \theta$}
                \STATE Generate predicate $p$: $p_k |\dots| p_j\quad$  
                \STATE Add $p$ to $\mathcal{P}$
                \STATE $k \gets j+1$
            \ELSE
                \STATE Generate predicate $p$: $ p_k |\dots| p_{|\mathcal{L}|}$
                \STATE Add $p$ to $\mathcal{P}$
                \STATE break
            \ENDIF
        \ENDIF
    \ENDFOR
  \RETURN $\mathcal{P}$
  \end{algorithmic}
\end{algorithm*}


\begin{algorithm*}
  \caption{Predicate generation for numeric features}
  \begin{algorithmic}[1]
  	\REQUIRE The dataset $\mathcal{D}$, the global predicate set $\mathcal{P}$, the minimum support threshold $\theta$
   \REQUIRE The categorical feature set $\mathcal{F}^c$, the numeric feature set $\mathcal{F}^n$
  	\STATE $\mathcal{T}\gets \emptyset$
  	\FOR{$i=1,\ldots,|\mathcal{F}^c|$}
        \STATE Train a DT classification model $DT(\mathcal{F}^n) \longrightarrow F^c_i$ on $\mathcal{D}$
        \STATE For each internal node $j$ in the trained DT model, add $(F^n_j,\tau_j, q_{\tau_j})$ to $\mathcal{T}$ based on its split rule
    \ENDFOR
    
    \FOR{$i=1,\ldots,|\mathcal{F}^n|$}
        \STATE Train a DT regression model $DT(\mathcal{F}^n_{-i}) \longrightarrow F^n_i$ on $\mathcal{D}$
        \STATE For each internal node $j$ in the trained DT model, add $(F^n_j,\tau_j, q_{\tau_j})$ to $\mathcal{T}$ based on its split rule 
    \ENDFOR
    
    \FOR{$i=1,\ldots,|\mathcal{F}^n|$}
        \STATE Get $F^n_i:(\tau_1, \dots,\tau_J)$ from $\mathcal{T}$ where $q_{\tau_1} \geq ...\geq q_{\tau_J}$
        \STATE $\mathcal{L}  \gets [ \tau_1 ]$ \COMMENT{list of cut-off values to keep}
        \FOR{$j=2,\ldots,J$}
            \STATE $k \gets 1$ \COMMENT{insert position of $\tau_j$}
            \WHILE{$k \leq |\mathcal{L}|$ and $\tau_j < \mathcal{L}[k]$ }
                \STATE $k \gets k+1$
            \ENDWHILE
            
            \IF{$k=1$}
                 \IF{$sup( \tau_j \leq F^n_i < \mathcal{L}[1]) > \theta$}
                    \STATE Insert $\tau_j$ to $\mathcal{L}$ at position $k$
                \ENDIF
            \ELSIF{$k=|\mathcal{L}|+1$}
                \IF{$sup( \mathcal{L}[k-1]\leq F^n_i < \tau_j ) > \theta$}
                    \STATE Insert $\tau_j$ to $\mathcal{L}$ at position $k$
                \ENDIF
            \ELSE
                \IF{$sup( \mathcal{L}[k-1]\leq F^n_i < \tau_j ) > \theta$ and $sup( \tau_j \leq F^n_i < \mathcal{L}[k]) > \theta$}
                    \STATE Insert $\tau_j$ to $\mathcal{L}$ at position $k$
                \ENDIF
            \ENDIF
        \ENDFOR
        \FOR{$j=1,\ldots,|\mathcal{L}|$}        
            \IF{$j=1$} 
                \STATE Generate predicate $p: F^n_i < \mathcal{L}[j]$ and add $p$ to $\mathcal{P}$
            \ENDIF
            \IF{$ 1<j \leq |\mathcal{L}|$}
                \STATE Generate predicate $p: \mathcal{L}[j-1] \leq F^n_i < \mathcal{L}[j]$ and add $p$ to $\mathcal{P}$
            \ENDIF
            \IF{$j=|\mathcal{L}|$}
                \STATE Generate predicate $p: F^n_i \geq \mathcal{L}[j]$ and add $p$ to $\mathcal{P}$
            \ENDIF
        \ENDFOR
    \ENDFOR
\RETURN $\mathcal{P}$
  \end{algorithmic}
   \label{alg:predicate_cont}
\end{algorithm*}

\subsection{Implementation Details for Experiments}
Regarding baseline anomaly explanation methods, we implement Anchor based on the Github repository in github.com/marcotcr/anchor. We set its hyperparameters $B=10$, $\epsilon=0.1$, $\delta=0.05$ as suggested in the original paper \cite{ribeiro2018anchors}. We implement LIME based on the Github repository in github.com/marcotcr/lime. We implement SHAP based on the Github repository in github.com/slundberg/shap. Specifically, the KernelExplainer is used with 20 samples generated by KMeans as background dataset for integrating out features. COIN and ATON are implementated based on the Github repository in github.com/xuhongzuo/outlier-interpretation. For all methods, the default hyperparameter values are used unless specifically mentioned here.

Regarding Isolation Forest (IF) and Autoencoder (AE), we implement IF using Scikit-Learn \cite{pedregosa2011scikit} library of version 1.1.2. The default hyperparameter values are used. We implement a vanilla version of AE with three hidden layers, the number of neurons in the bottleneck layer is set to $\frac{1}{3}$ of the input dimension. Each model is trained with 10 epochs to minimize the reconstruction error on the training set. All of our experiments were run on a Linux machine with 64 GiB memory, 8 4.2GHz Intel Cores and a GTX 1080 GPU.

\subsection{Further Experiments for Ablation Study}
 We also demonstrate the importance of our dependency-based predicate generation method for numeric features by conducting an ablation study. Concretely, we replace our dependency-based predicate generation method for numeric features with an uniform interval-based and a KMeans-based discretization method which simply discretize the value of numeric features to 10 bins for predicate generation. Then, we generate PARs for the benchmark datasets based on three different predicate generation method for numeric features, namely Uniform Bins, KMeans Bins and dependency-based. Furthermore, we compare two important metrics on the performance of anomaly explanation: PoF@TPs, the average accuracy score of the top1 PAR. The result is given in Table~\ref{tab:bin_comp}. As can be seen, there are discernible improvements on both metrics when using our dependency-based method for predicate generation of numeric features compared with Uniform Bins and KMeans Bins methods.

\begin{table*}
  \caption{PoF@TPs and the average accuracy score of the top1 PAR, when using different methods to generate predicates for numeric features.}
  \label{tab:bin_comp}
  \centering
  \begin{tabular}{lccccccccc}
    \toprule
    \multirow{2}{*}{Dataset} & \multicolumn{2}{c}{Uniform Bins}& \multicolumn{2}{c}{KMeans Bins} & \multicolumn{2}{c}{Dependency-based} \\
    \cmidrule(r){2-3} \cmidrule(r){4-5} \cmidrule(r){6-7} 
    & PoF@TPs & Accuracy score & PoF@TPs & Accuracy score & PoF@TPs & Accuracy score\\
    \midrule
    breastw & 1.00 & 5.54 & 1.00 & 5.58 & 1.00 & 5.69 \\
cardio & 1.00 & 5.49 & 1.00 & 5.43 & 1.00 & 5.73 \\
Cardiotocography & 0.98 & 5.39 & 1.00 & 5.20 & 1.00 & 5.45 \\
fault & 0.99 & 5.02 & 1.00 & 5.11 & 1.00 & 5.23 \\
Ionosphere & 1.00 & 5.97 & 1.00 & 6.00 & 1.00 & 6.00 \\
Lymphography & 1.00 & 6.00 & 1.00 & 6.00 & 1.00 & 6.00 \\
magic & 0.63 & 4.32 & 0.64 & 3.46 & 0.79 & 4.32 \\
Pima & 0.31 & 3.20 & 0.27 & 2.40 & 0.63 & 4.13 \\
satellite & 1.00 & 5.57 & 0.87 & 5.63 & 1.00 & 5.67 \\
satimage-2 & 1.00 & 5.88 & 1.00 & 6.00 & 1.00 & 6.00 \\
shuttle & 0.99 & 5.98 & 1.00 & 5.99 & 1.00 & 5.99 \\
skin & 0.22 & 1.65 & 0.36 & 1.68 & 0.43 & 2.32 \\
Stamps & 0.27 & 1.44 & 0.29 & 5.09 & 0.86 & 5.28 \\
thyroid & 0.72 & 5.38 & 0.85 & 5.45 & 0.82 & 5.49 \\
WBC & 1.00 & 6.00 & 1.00 & 6.00 & 1.00 & 6.00 \\
WDBC & 1.00 & 6.00 & 1.00 & 6.00 & 1.00 & 6.00 \\
wine & 1.00 & 6.00 & 1.00 & 6.00 & 1.00 & 6.00 \\
\midrule
Avg. & 0.83 & 4.99 & 0.84 & 5.12 & \textbf{0.91} & \textbf{5.37} \\
\bottomrule
  \end{tabular}
\end{table*}

\subsection{Detailed Experiment Results}
We give the dataset-specific results for Table 4 of the main paper in Table~\ref{tab:rule_comp}. We give the dataset-specific results for Table 5 of the main paper in Table~\ref{tab:afi_comp}. The dataset-specific results for PoF@TPs and PoF@FPs are given in Table~\ref{tab:tp_fp_fpr}.
\begin{table*}
  \caption{The average time cost to compute the anchors and top 5 PARs at each anomaly instance in the benchmark datasets identified by IF and AE models, and the average precision, recall and F1 score of using anchors and top 5 PARs computed at each anomaly instance to do anomaly detection in unseen data.}
  \label{tab:rule_comp}
  \centering
  \begin{tabular}{lcccccccc}
    \toprule
    \multirow{2}{*}{Dataset} & \multicolumn{2}{c}{ Precision}& \multicolumn{2}{c}{ Recall} & \multicolumn{2}{c}{F$_1$ score} & \multicolumn{2}{c}{Time Cost (secs)}   \\
    \cmidrule(r){2-3} \cmidrule(r){4-5} \cmidrule(r){6-7} \cmidrule(r){8-9}
    & Anchor & PAR& Anchor & PAR& Anchor & PAR & Anchor & PAR\\
    \midrule
    breastw & 0.97 & 0.83 &  0.45 & 0.58 &  0.59 & 0.65 & 5.67 & 0.03 \\
    cardio & 0.29 &  0.61 & 0.03 & 0.16 & 0.05 & 0.25 &  35.34 &0.27 \\
    Cardiotocography & 0.54 & 0.55 & 0.03 & 0.05 & 0.05 & 0.09 & 16.89 & 0.35 \\
    fault & 0.43 & 0.48 &  0.07 & 0.02 & 0.12 & 0.03 & 3.19 & 1.77 \\
    Ionosphere & 0.61 & 0.94 & 0.03 & 0.50 & 0.05 & 0.64 & 79.69 &0.11 \\
    Lymphography & 1.00 & 0.61 & 0.33 & 0.56 & 0.50 & 0.57 & 6.72 & 0.10 \\
    magic & 0.69 & 0.68 & 0.07 & 0.04 & 0.12 & 0.07 & 3.59 & 0.10 \\
    Pima & 0.47 & 0.44 & 0.04 & 0.04 & 0.07 & 0.07 & 4.19 & 0.03 \\
    satellite & 0.88 & 0.87 & 0.05 & 0.19 & 0.08 & 0.29 & 56.86 & 0.43 \\
    satimage-2 & 0.49 & 0.85 & 0.42 & 0.44 & 0.32 & 0.51 & 88.64 & 0.26 \\
    shuttle & 0.93 & 0.99 & 0.17 & 0.95 & 0.24 & 0.97 & 5.97 & 0.04 \\
    skin & 0.52 & 0.57 & 0.41 & 0.19 & 0.43 & 0.28 & 1.53 & 0.01 \\
    Stamps & 0.50 & 0.33 & 0.22 & 0.20 & 0.30 & 0.25 & 4.52 &0.03 \\
    thyroid & 0.08 & 0.65 & 0.26 & 0.17 & 0.09 & 0.27 & 3.77 & 0.02 \\
    WBC & 0.50 & 0.59 & 0.67 & 0.92 & 0.57 & 0.71 & 6.38 & 0.02 \\
    WDBC & 0.00 & 0.72 & 0.00 & 1.00 & 0.00 & 0.84 & 108.9 & 0.09 \\
    wine & 0.50 & 0.67 & 0.24 & 0.67 & 0.33 & 0.67 & 5.16 & 0.03 \\
    \midrule
    Avg. & 0.55 &  0.67 &  0.21 &  0.39 & 0.23 & 0.42 & 25.71 & 0.22 \\
    Improvement. & - &  +22\% &  - &+86\% & - & +83\% & - & -99\% \\
    \bottomrule
  \end{tabular}
\end{table*}

\begin{table*}
  \caption{Accuracy of abnormal feature identification in terms of HitRate@100\% and HitRate@150\%.}
  \label{tab:afi_comp}
  \centering
  \begin{tabular}{lcccccc}
    \toprule
    Dataset & LIME & SHAP  & COIN & ATON & PAR \\
    \midrule
    breastw & 0.19/0.26 & 0.65/0.87 &  0.48/0.56 &  0.80/0.80  &  0.69/0.69 \\
    cardio & 0.38/0.55 & 0.73/0.78 &  0.42/0.49 &  0.54/0.64  &  0.60/0.68 \\
    Cardiotocography & 0.31/0.47 & 0.65/0.73 &  0.38/0.50 &  0.68/0.72  &  0.76/0.76 \\
    fault & 0.17/0.25 & 0.42/0.58 &  0.33/0.55 &  0.52/0.58  &  0.75/0.75 \\
    Ionosphere & 0.43/0.43 & 0.60/0.60 &  0.43/0.43 &  0.51/0.66  &  0.63/0.70 \\
    Lymphography & 0.33/0.50 & 0.17/0.33 &  0.65/0.88 &  0.48/0.68  &  0.83/1.00\\
    magic & 0.48/0.58 & 0.70/0.82 &  0.64/0.74 &  0.72/0.80  &  0.69/0.77\\
    Pima & 0.61/0.71 & 0.86/0.91 &  0.66/0.78 &  0.64/0.88  &  0.67/0.76\\
    satellite & 0.08/0.11 & 0.21/0.30 &  0.66/0.68 & 0.48/0.56  &  0.67/0.68\\
    satimage-2& 0.04/0.04 & 0.16/0.20 &  0.48/0.57   &  0.50/0.52&  0.56/0.56\\
    shuttle & 0.41/0.53 & 0.75/0.81 & 0.67/0.73 &  0.60/0.68  &  0.62/0.67 \\
    skin & 0.74/0.92 & 0.90/0.95 & 0.85/0.98 &  0.84/0.98 & 0.84/0.84 \\
    Stamps & 0.58/0.64 & 0.64/0.78 & 0.54/0.64 &  0.58/0.68 & 0.61/0.61 \\
    thyroid & 0.71/0.82 & 0.90/0.95 & 0.60/0.71 &  0.64/0.74 & 0.72/0.75 \\
    WBC & 0.37/0.50 & 0.73/1.00 & 0.66/0.83 &  0.88/0.98 & 0.53/0.53 \\
    WDBC& 0.04/0.04 & 0.17/0.17 & 0.16/0.33 &  0.42/0.42 & 0.42/0.42 \\
    wine& 0.69/0.74 & 0.81/0.85 &  0.54/0.68 &  0.65/0.71  &  0.57/0.79 \\
    \midrule
    Avg.& 0.39/0.48 & 0.59/0.68 &  0.54/0.65 &  0.62/\textbf{0.71}  &  \textbf{0.66}/0.70 \\
    Avg. Rank & 4.47/4.35 & 2.24/\textbf{2.12} & 3.53/3.18 & 2.53/2.47 & \textbf{2.00}/2.53 \\
    \bottomrule
  \end{tabular}
\end{table*}

\begin{table*}
  \caption{The PoF for explaining predicated anomalies when the predicated anomalies are True Positives (TPs) and False Positives (FPs). When there is no FP, we set PoF to n.a.}
  \label{tab:tp_fp_fpr}
  \centering
  \begin{tabular}{lccc}
    \toprule
    \multirow{2}{*}{Dataset} & Num. & PoF & PoF \\
    & features & @TPs & @FPs  \\
    \midrule
    breastw & 9 & 1.00 & 1.00\\ 
    cardio & 21 & 1.00 & 1.00 \\
    Cardiotocography & 21 & 1.00 & 0.94 \\ 
    fault & 27 & 1.00 & 0.97 \\
    Ionosphere & 32 & 1.00 & 1.00 \\
    Lymphography & 18 & 1.00 & n.a \\
    magic & 10 & 0.79 & 0.23 \\
    Pima & 8 & 0.63 & 0.27 \\
    satellite & 36 & 1.00 & 0.87 \\
    satimage-2 & 36 & 1.00 & n.a \\
    shuttle & 9 & 1.00 & 0.55 \\
    skin & 3 & 0.43 & 0.32 \\
    Stamps & 9 & 0.86 & 0.80 \\
    thyroid & 6 & 0.82 & 0.25 \\
    WBC & 9 & 1.00 & n.a \\
    WDBC & 30 & 1.00 & 1.00 \\
    wine & 13 & 1.00 & 1.00 \\
    \midrule
    Avg. & 17.47 & 0.91 & 0.73  \\
    \bottomrule
  \end{tabular}
\end{table*}


\end{document}